\documentclass{article}

\usepackage{booktabs}
\usepackage{titlesec}

\usepackage{multirow}

\usepackage{graphicx}
\usepackage{accsupp}

\usepackage{PRIMEarxiv}

\usepackage[utf8]{inputenc} 
\usepackage[T1]{fontenc}    
\usepackage{hyperref}       
\usepackage{url}            
\usepackage{booktabs}       
\usepackage{amsfonts}       
\usepackage{nicefrac}       
\usepackage{microtype}      
\usepackage{lipsum}
\usepackage{fancyhdr}       
\usepackage{graphicx}       
\graphicspath{{media/}}     
\usepackage{amsmath}

\usepackage{authblk}

\begin{document}

\title{CAAT-EHR: Cross-Attentional Autoregressive Transformer for Multimodal Electronic Health Record Embeddings
}

   \author[1, 2, 3]{\small Mohammad Al Olaimat}
\author[1, 2, 3]{\small Shaika Chowdhury}
\author[1, 2, 3, 4$\ast$]{\small Serdar Bozdag}

\author[5]{\small for the Alzheimer’s Disease Neuroimaging Initiative}


  \affil[1]{\footnotesize Department of Computer Science and Engineering, University of North Texas, Denton, TX 76203, USA}
\affil[2]{BioDiscovery Institute, University of North Texas, Denton, TX 76203, USA}
\affil[3]{Center for Computational Life Sciences, University of North Texas, Denton, TX 76203, USA}
\affil[4]{Department of Mathematics, University of North Texas, Denton, TX 76203, USA}
\affil[5]{Data used in
preparation of this article were obtained from the Alzheimer’s Disease Neuroimaging Initiative (ADNI) database (adni.loni.usc.edu). As such, the investigators within the ADNI contributed to the design and implementation of ADNI and/or provided data but did not participate in the analysis or writing of this report. A complete listing of ADNI investigators can be found at:
http://adni.loni.usc.edu/wp-content/uploads/how\_to\_apply/ADNI\_Acknowledgement\_List.pdf}
\affil[$\ast$]{Corresponding author Serdar.Bozdag@unt.edu}

\maketitle

\begin{abstract}
\textbf{Motivation:} Electronic Health Records (EHRs) contain rich, longitudinal patient information across structured (e.g., labs, vitals, and imaging) and unstructured (e.g., clinical notes) modalities. While deep learning models such as RNNs and Transformers have advanced single- and multimodal EHR analysis, existing methods often optimize for specific downstream tasks and overlook the creation of generalizable patient representations that can be reused across multiple tasks. To address this gap, we propose \textbf{CAAT-EHR}, a novel Cross-Attentional Autoregressive Transformer architecture that produces task-agnostic, longitudinal embeddings of multimodal EHR data. In CAAT-EHR, self-attention layers capture temporal dependencies within each modality, while cross-attention layers fuse information across modalities to model complex interrelationships. During pre-training, an autoregressive decoder predicts future time steps from the fused embeddings, enforcing temporal consistency and enriching the encoder output. Once trained, the encoder alone generates versatile multimodal EHR embeddings that can be applied directly to a variety of predictive tasks. 
\\\textbf{Results:} CAAT-EHR demonstrates significant improvements on benchmark EHR datasets for mortality prediction, ICU length-of-stay estimation, and Alzheimer’s disease diagnosis prediction. Models using EHR embeddings generated by CAAT-EHR outperform models trained on raw EHR data in eleven out of twelve comparisons for F1 score and AUC across all three downstream tasks. Ablation studies confirm the critical roles of cross-modality fusion and autoregressive refinement. Overall, CAAT-EHR provides a unified framework for learning generalizable, temporally consistent multimodal EHR representations that support more reliable clinical decision support systems. 
\\\textbf{Availability and Implementation:} The source code and documentation of CAAT-EHR are available at \url{https://github.com/bozdaglab/CAAT-EHR}.
\\\textbf{Contact:} \href{mailto:serdar.bozdag@unt.edu}{Serdar.Bozdag@unt.edu}
\\\textbf{Supplementary:} Uploaded as an attachment.
\end{abstract}


\section{Introduction}
The increasing adoption of Electronic Health Records (EHRs) has resulted in the accumulation of a vast amount of longitudinal patient data. These data, which span multiple modalities, such as structured data (e.g., lab results, imaging, and vital signs) and unstructured data (e.g., clinical notes), provide a comprehensive, yet complex view of patient health \cite{hossain2023nlp}. EHRs have emerged as a fundamental resource for modeling patient diagnoses and classifications, as well as disease progression and subtyping. By leveraging advanced techniques such as statistical approaches, machine learning, and deep learning (DL), they enable healthcare providers to process large volumes of data, extract valuable insights, and make accurate, data-driven clinical decisions \cite{zhang2019attain,li2019early,tan2021cooperative,edduweh2024liouville}.

The effective integration and representation of these data are critical for predictive modeling tasks, such as mortality prediction, disease diagnosis prediction, and length-of-hospital-stay estimation. However, traditional machine learning models like Random Forest (RF) and Support Vector Machine (SVM) often fail to capture the temporal and multimodal dependencies in such datasets, as they typically rely on a single time point, such as baseline or the latest visit. Alternatively, decisions can be made on aggregated data across all time points; however, this approach often oversimplifies the data by ignoring temporal dynamics and relationships between modalities, potentially leading to suboptimal performance.

Recurrent neural networks (RNN), such as Long Short-Term Memory (LSTM) \cite{hochreiter1997lstm} and Gated Recurrent Unit (GRU) \cite{cho2014properties}, and Transformer \cite{vaswani2017attention} architectures, originally designed for natural language processing (NLP), have emerged as powerful tools for modeling sequential data. Their ability to capture long-range dependencies and contextual relationships makes them particularly well-suited for EHR data.

The analysis of EHRs has undergone a transformative evolution with the use of RNN and Transformers. Early efforts predominantly focused on modeling a single data modality. Methods such as RETAIN \cite{choi2016retain}, T-LSTM \cite{baytas2017patient}, DATA-GRU \cite{tan2020datagru}, EHR2Vec \cite{wang2020ehr2vec}, BiCMT \cite{wang2021bicmts}, CJANe \cite{tan2021cooperative}, and KIT-LSTM \cite{liu2022kitlstm} 
employed RNN or Transformers to derive latent representations of sequential EHR data. These representations were trained and evaluated on the same task, such as mortality prediction or disease progression. While effective for single-modality data, such methods are limited in two ways: they are optimized for a specific task, which hinders their ability to generalize across diverse downstream tasks, and they lack the capability to capture the complexities of multimodal data, which is often crucial in EHR analysis.

To address the multifaceted nature of EHR data, researchers shifted toward multimodal EHR analysis. A foundational approach involved early integration, where data from various modalities (e.g., clinical notes, lab tests, imaging, and diagnoses) were concatenated into a single sequence and processed by RNN such as \cite{nguyen2020adprogression}, PPAD \cite{al2023ppad}, and TA-RNN \cite{al2024ta}. Latent representations of the concatenated data were pooled and fed into a multi-layer perceptron (MLP) for predictions. Despite their simplicity, early integration approaches often failed to exploit the unique characteristics of each modality or their intricate interrelationships.

Subsequent advances explored separate processing of modalities, with studies such as \cite{lee2019multimodal,lyu2022multimodal, hayat2022medfuse, yoon2023ehrs} employing distinct RNN or Transformer for each modality. These models generated modality-specific latent representations, which were later concatenated into a unified vector for downstream tasks. While this paradigm preserved modality-specific temporal dynamics, it struggled to model inter-modality interactions effectively, limiting its ability to capture the full complexity of patient trajectories.


To address these gaps, MedFuseNet \cite{sharma2021medfusenet}, MADDi \cite{golovanevsky2022multimodal}, and TransformEHR \cite{yang2023transformehr} use self-attention and cross-attention to model intra- and inter-modality relationships. While effective, these models often face scalability challenges due to the computational overhead associated with attending across and within all modalities.

To overcome this limitation, the One-Versus-Others (OvO) attention model \cite{golovanevsky2024onevsothers} introduced a scalable multimodal integration approach where each modality attends to the averaged information from all other modalities, excluding itself. This design captures cross-modal interactions efficiently without relying on self-attention within each modality, thereby reducing computational cost and improving scalability.

The introduction of pre-training paradigms, such as BEHRT \cite{li2020behrt} and Med-BERT \cite{rasmy2021medbert} marked a significant milestone in EHR modeling. Drawing inspiration from NLP models like BERT \cite{devlin2018bert}, BEHRT utilizes a masked language modeling (MLM) objective to pre-train contextual representations of EHR sequences. These pre-trained models are later fine-tuned on specific downstream tasks, achieving state-of-the-art results. However, BEHRT and Med-BERT primarily focused on textual data (e.g., clinical notes) and did not incorporate mechanisms for effective integration of multimodal EHR data.
Despite advancements, existing methodologies often oversimplify multimodal integration or fail to capture intricate interdependencies across modalities. Additionally, many models focus on task-specific optimization, overlooking the need for generalized longitudinal representations of EHR data transferable to a wide range of downstream tasks. These limitations highlight the need for a holistic solution that integrates multimodal data effectively while generating versatile and temporally consistent embeddings.

To address these gaps, in this study, we propose CAAT-EHR: Cross-Attentional Autoregressive Transformer for Multimodal Electronic Health Record Embeddings, a novel architecture designed to advance EHR modeling by generating robust task-agnostic longitudinal representations of EHR data. At the core of CAAT-EHR there is an encoder, which generates embeddings that capture the temporal and contextual relationships inherent in the data. When EHR data consist of  multiple modalities, the encoder employs cross-attention mechanisms to facilitate the integration of these modalities. This enables the model to comprehensively capture interactions and dependencies across modalities, resulting in enriched and holistic embeddings. To refine and optimize these embeddings further, a decoder operates as an autoregressive module, predicting future data from the processed sequence. 

CAAT-EHR addresses the limitations of prior methods through the following innovations, which together form a unique approach to EHR representation learning:\begin{itemize}
\item \textbf{Cross-Attention Mechanisms for Multimodal Fusion:} 
CAAT-EHR  integrates self- and cross-attention mechanisms within a single framework for task-agnostic longitudinal representation learning. In this model, self-attention is employed to capture temporal and contextual relationships within each modality, enabling the model to learn rich intra-modality representations. In addition, cross-attention is used to integrate information across multiple modalities. This enables a comprehensive fusion of multimodal data and generates enriched embeddings that represent complex interdependencies between modalities.
\item \textbf{Task-Agnostic Longitudinal Representation:} Task-agnostic longitudinal embeddings, while explored in prior work (e.g., BEHRT for textual data), have not been applied in a unified framework for EHR data. To the best of our knowledge, CAAT-EHR is the first to combine these representations with self- and cross-attention mechanisms to capture the temporal evolution of patient data in a way that is independent of specific downstream tasks. 
\item \textbf{Autoregressive Temporal Refinement:} During pre-training, the autoregressive decoder enhances the encoder's longitudinal embeddings by predicting future data points, ensuring temporal consistency and alignment. The decoder acts solely as a supervision mechanism to optimize the encoder’s output. After pre-training, only the encoder is used to generate task-agnostic longitudinal embeddings applicable to various downstream tasks.

\item \textbf{Unified Multimodal Integration:}
This study advances multimodal clinical modeling by integrating structured, unstructured, and/or omics data within the EHR, addressing a notable gap in existing work.
\end{itemize}

Extensive evaluations on benchmark datasets demonstrated that models trained on the embeddings generated by CAAT-EHR outperformed those trained on raw data and baseline embeddings in diverse tasks such as mortality prediction,  length of intensive care unit (ICU) stay prediction, and Alzheimer’s disease (AD) diagnosis prediction. Ablation studies highlighted the role of cross-attention in multimodal fusion and the autoregressive decoder in refining temporal consistency.

\section{Materials and Methods}\label{sec2}

\subsection{Datasets}\label{subsec1}

In the following subsections, we introduce each dataset used to evaluate the proposed model, describe the preprocessing steps performed, and give key statistics.

\subsubsection{The MIMIC-III Dataset}
The MIMIC-III database \cite{johnson2016mimic,goldberger2000physionet} is a comprehensive repository of EHRs that contains records of patients in the ICU. Following the procedures outlined in \cite{harutyunyan2019multitask}, we extracted a subset of patient time-series data from the MIMIC-III database. This dataset has 17 clinical features for 1,730,641 time points from 18,094 patients. Since patients in MIMIC-III can have more than one ICU stay, the analysis was conducted at the ICU stay level, with each stay considered a separate sample rather than aggregating at the patient level. The dataset comprises 21,139 unique ICU stays. Each ICU stay represents time-series EHR data containing between two and 2,879 time points, depending on data availability and recording frequency. Each time point corresponds to a specific timestamp in hours, minutes, and seconds, from which the intervals between consecutive time points can be calculated. These intervals vary both within a patient and across patients, ranging from one to 2,040 minutes. 
Of the 17 features, 12 belong to the \textit{continuous data modality}, while five belong to the \textit{categorical data modality}. The selected features and their data modalities are listed in Supplemental Table 1.


The dataset has a high proportion of missing values (Supplemental Table 1) and variability in the number of time points and intervals. To address these challenges, the dataset underwent several preprocessing steps. 
ICU stays were excluded from the dataset if they had fewer than three time points or if at least one feature was completely missing (i.e., never collected during the stay). A total of 252 ICU stays were eliminated, comprising 231 stays due to missing data and 21 stays with fewer than three time points. Following the approach described in \cite{harutyunyan2019multitask}, missing values were then imputed using the most recent available measurement when present. If no prior measurement was available for a missing value (e.g., when a feature's first recorded value occurs only after one or more missing values), the missing value was replaced with a predefined 'normal' value, selected from the set of possible valid values for the feature (Supplemental Table 2). For categorical features, the possible values and their meanings are detailed in Supplemental Table 3, based on \cite{bodien2021gcs,monteerarat2022capillary}. 

In addition to continuous and categorical features, we used MIMIC-III \textit{clinical notes} as a third data modality for the same ICU stays. 
Across all stays, there were 37 notes per ICU stay on average, with a mode of 12. Within each ICU stay, notes were sorted according to their timestamp, which is recorded in YYYY-MM-DD HH:MM:SS format. The clinical notes were aligned with the categorical and continuous modalities using the hourly time differences of the latter. Note timestamps were computed from the time differences between consecutive notes and mapped to the nearest preceding hour. Each hour was assigned the embedding of the most recent clinical note available at that time. Embeddings were then forward-filled between note events to ensure temporal consistency. We applied TF-IDF to generate an initial note representation with a feature dimension of 30. We intentionally avoided more sophisticated note representation methods to enable CAAT-EHR to demonstrate its ability to generate expressive representations for each data modality.  

The final dataset contains 20,887 ICU stays, with an average of 82.04 time points per stay (Supplemental Figure 1). To reduce the dataset size, we limited each stay to the most recent 200 time points, thereby avoiding the need to pad stays to the maximum length of 2,879 time points. There were only 35 stays that had more than 200 time points, thus this trimming had minimal effect on the dataset. For ICU stays containing fewer than 200 time points, remaining time points were padded with a value of –50. One-hot encoding was applied to the categorical features, resulting in 30 features derived from the five categorical features. 

The data was then divided into two subsets: (1) the MIMIC-III embedding task dataset, comprising 70\% of the data, which was used for pre-training CAAT-EHR to learn generalizable task-agnostic longitudinal representations from longitudinal EHR data, capturing the temporal dynamics and dependencies across time points; and (2) the MIMIC-III downstream task dataset, comprising 30\% of the data, which was used for downstream mortality and length-of-stay prediction tasks. Finally, both the embedding and downstream task datasets were independently normalized using feature-wise z-normalization for the continuous data modality. Since patients in the dataset may have more than one ICU stay, to avoid any potential data leakage, we ensured that ICU stays belonging to the same patient were not included in both the embedding and downstream task datasets, or both in train and test splits in the embedding or downstream task datasets. 


\subsubsection{The ADNI Dataset}
The ADNI database \url{https://adni.loni.usc.edu/} provides longitudinal data aimed at advancing research in AD and related conditions. The ADNImerge R package (available at \url{https://adni.bitbucket.io/}) was used to extract a subset of time-series patient data for 15,087 clinical time points from 2,288 individuals from the ADNI database, along with diagnoses at each time point. Several preprocessing steps were performed, following the procedures outlined in PPAD \cite{al2023ppad} such as eliminating features and samples with high missing rate, feature imputation, and feature normalization.  
Unlike PPAD, we also utilized cognitively normal (CN) cases for pre-training. After preprocessing, the final dataset consisted of 19 longitudinal features (12 related to \textit{cognitive performance} and seven to \textit{MRI data}) from 1,296 patients across 6,096 time points, as detailed in Supplemental Table 4. 

We further incorporated \textit{DNA methylation} as a third modality for the same patients, measured at three time steps. This data was chronologically aligned with the cognitive and MRI data at corresponding time points. We performed Singular Value Decomposition (SVD) on the raw DNA methylation data to reduce its original dimension from 116,227 to 64. The dataset was then divided into two subsets: the ADNI embedding task dataset, comprising 40\% of the data and used to pre-train CAAT-EHR, and the ADNI downstream task dataset, comprising 60\% of the data and used for the downstream AD diagnosis prediction task.

Note that we used different split proportions based on dataset size: the smaller ADNI dataset used a 40/60 pre-training–downstream split, while the larger MIMIC-III dataset used a 70/30 split since 30\% was sufficient for downstream evaluation.


\subsubsection{Dataset Notations}

Let $\mathcal{M}$ denote the input data consisting of $K$ modalities, 
$\mathcal{M} = \{M_1, M_2, \ldots, M_K\}$, where $K$ = 3 in this study. Each modality $\mathcal{M}_i$ corresponds to a longitudinal EHR data modality (e.g., continuous, categorical, notes, or DNA methylation) with its own feature dimension $d_i$ and is associated with $N$ samples (e.g., patients), where $\mathcal{M}_i = \{X_1, X_2, \ldots, X_N\}$. 
Each sample $X$ represents measurements of $D$ features collected over $T$ time points from all data modalities. Mathematically 
$X = \{x_1, x_2, \ldots, x_T\} \in \mathbb{R}^{T \times D}$ where $D=\sum_{i=1}^{K}d_i$.

For each time point $t \in \{1, 2, \ldots, T\}$, $
x_t = \{x_t^1, x_t^2, \ldots, x_t^D\} \in \mathbb{R}^D
$
represents a vector of features of sample $X$ at time point $t$.

For each feature $f \in \{1, 2, \ldots, D\}$, 
$x^f = \{x_1^f, x_2^f, \ldots, x_T^f\} \in \mathbb{R}^T$ represents the $f$-th feature value of sample $X$ across all $T$ time points. Similarly, $x_t^f$ represents the $f$-th feature value of sample $X$ at time point $t$.

Finally, each sample $X$ is associated with a corresponding label $y$. In this study, $y \in \{0, 1\}$, where:
\begin{itemize}
    \item $y = 0$ denotes MCI (Mild Cognitive Impairment) and $y = 1$ denotes AD (Alzheimer’s Disease) for the AD prediction task.
    \item $y = 0$ denotes the absence of mortality and $y = 1$ denotes mortality in the mortality prediction task.
    \item $y = 0$ denotes short length-of-stay and $y = 1$ denotes long length-of-stay in the ICU length-of-stay prediction task.
\end{itemize}

It is important to emphasize that the label $y$ is used exclusively in the downstream task and not during the pre-training of CAAT-EHR.

\subsubsection{The Embedding Task Data}
After data preprocessing, both the MIMIC-III and ADNI datasets were divided into two parts: embedding task data and downstream task data.

In this study, the entire embedding task data were exclusively used for pre-training CAAT-EHR to learn generalizable task-agnostic longitudinal representations from longitudinal EHR data. The downstream task data were used after the pre-training of CAAT-EHR to evaluate CAAT-EHR’s ability to generate task-agnostic representations of longitudinal EHR data. 

For pre-training, the embedding task dataset was partitioned into input features and prediction targets. For time-series EHR data with $T$ time points, data from the first time point up to $T-2$ were used as input features, while data from $T-1$ to the last time point $T$ were used as prediction targets for the model to learn. In other words, during pre-training, CAAT-EHR was trained using the input features from the training portion of the dataset up to the last two time points to predict the corresponding feature values in the last two time points in the dataset.

\subsubsection{The Downstream Task Data}\label{sec:downstream-task-data}
\begin{figure}[htbp]
    \centering
  \BeginAccSupp{method=pdfstringdef,ActualText={Diagram illustrating the CAAT-EHR data processing and modeling workflow. (A) Raw EHR data are preprocessed to construct datasets for embedding learning and downstream tasks. (B) The CAAT-EHR model is pre-trained on the embedding task dataset. (C) The pre-trained encoder generates task-agnostic longitudinal embeddings from the downstream task dataset, which are used for prediction tasks including mortality status, ICU length of stay (MIMIC-III), and Alzheimer’s disease diagnosis (ADNI).}}\includegraphics[width=1\linewidth]{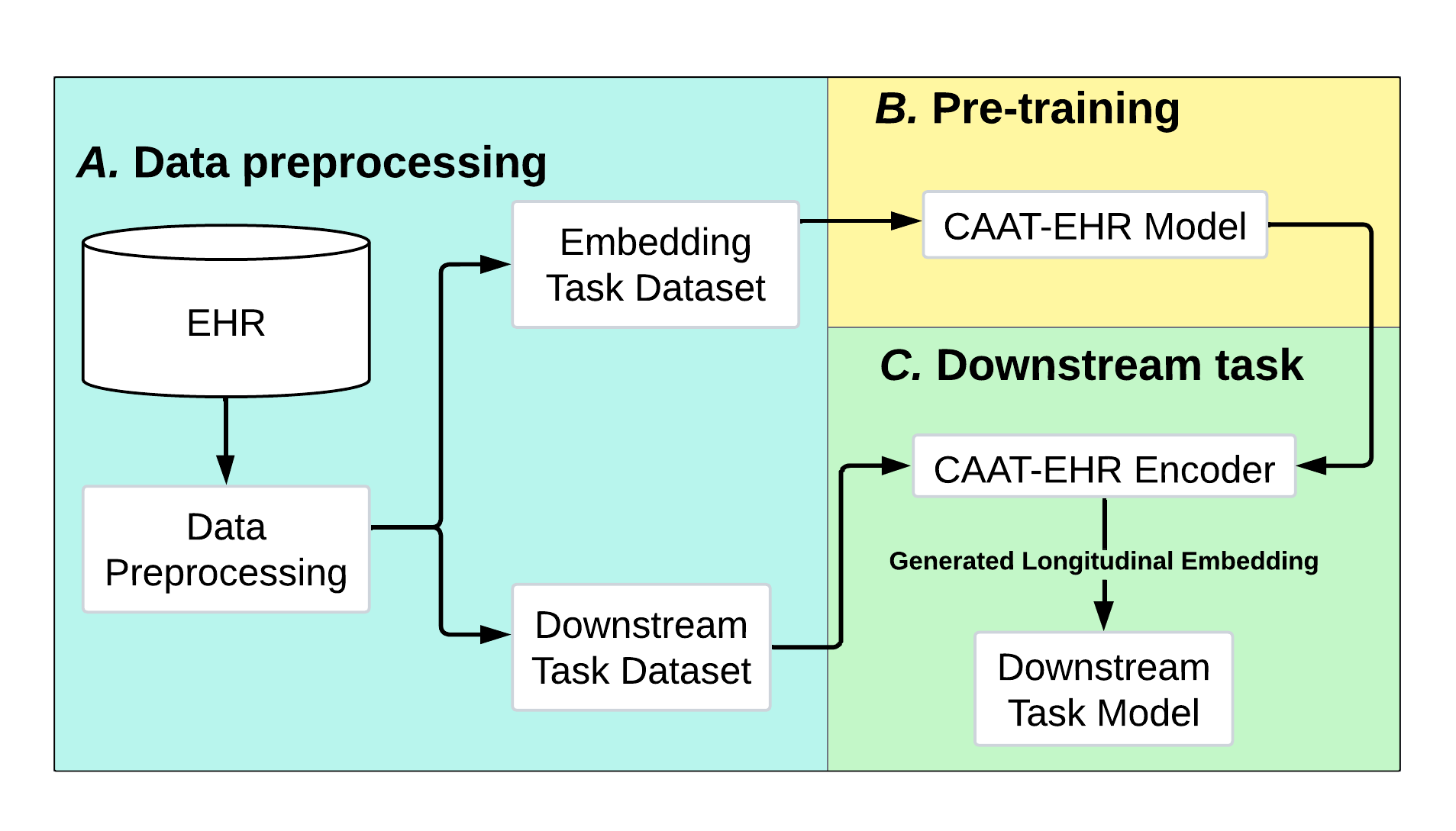}
  \EndAccSupp{}
    \caption{Overview of data processing and model workflow: (A) EHR data preprocessing to create embedding and downstream task datasets. (B) Pre-training CAAT-EHR using the embedding task dataset. (C) Generating task-agnostic longitudinal embeddings from the downstream task dataset using the pre-trained encoder for prediction tasks, namely mortality status, ICU length of stay (MIMIC-III), and AD diagnosis (ADNI).}
    \label{fig:processing}
\end{figure}

To evaluate whether pre-trained CAAT-EHR generates task-agnostic embeddings from raw EHR data, we evaluated it for several downstream tasks. For these experiments, we utilized the downstream task data. Specifically, the pre-trained encoder of CAAT-EHR was retained and applied to generate enhanced generalizable longitudinal embeddings from the raw downstream task datasets. These embeddings were then utilized to train predictive models for the downstream tasks. The utilization of the embedding and downstream task datasets is illustrated in Figure~\ref{fig:processing}.

For the MIMIC-III dataset, there were two prediction tasks: mortality status (i.e., dead or alive) of an ICU patient and length of ICU stay. For each ICU patient, the $T-2$ sequence of time points was used as input features, and the target label represented either mortality status or length of stay. We excluded the final two time points to prevent potential label leakage. The length of stay, originally measured in days, was converted into a binary classification problem: stays of seven days or fewer were categorized as \emph{short}, while stays longer than seven days were categorized as \emph{long}. 

For the ADNI data, the prediction task was to predict AD diagnosis at the last time point. Thus, for each patient, data from the first clinical time point up to the second from the last time point were used as input features, while the diagnosis label at the last clinical time point was used as the target label.

\subsection{The Proposed Method}
In this study, we propose CAAT-EHR: Cross-Attentional Autoregressive Transformer for Multimodal Electronic Health Record Embeddings, designed to effectively model EHR data, particularly when it spans multiple modalities. The architecture is composed of two primary components: an encoder that generates task-agnostic longitudinal embeddings for the raw EHR data by leveraging both self- and cross-attention mechanisms, and a decoder that acts solely as a supervision mechanism to optimize the encoder’s output through autoregressive modeling in the pre-training stage. These embeddings incorporate information not only from the raw features but also from their temporal dynamics and contextual relationships within and across data modalities. This design ensures the effective representation of temporal data and dependencies, enabling robust embeddings that are suitable for various downstream tasks. Importantly, after pre-training, only the encoder of CAAT-EHR is retained to generate task-agnostic longitudinal embeddings that capture both the intrinsic characteristics of the data and the interactions across modalities for application in various downstream tasks.

\begin{figure}[htbp]
    \centering  \BeginAccSupp{method=pdfstringdef,ActualText={Schematic of the CAAT-EHR architecture showing a cross-attentional autoregressive Transformer that integrates multimodal longitudinal EHR inputs, applies cross-attention for modality fusion, and uses an autoregressive decoder to generate task-agnostic patient embeddings.}}\includegraphics[width=1.0\linewidth]{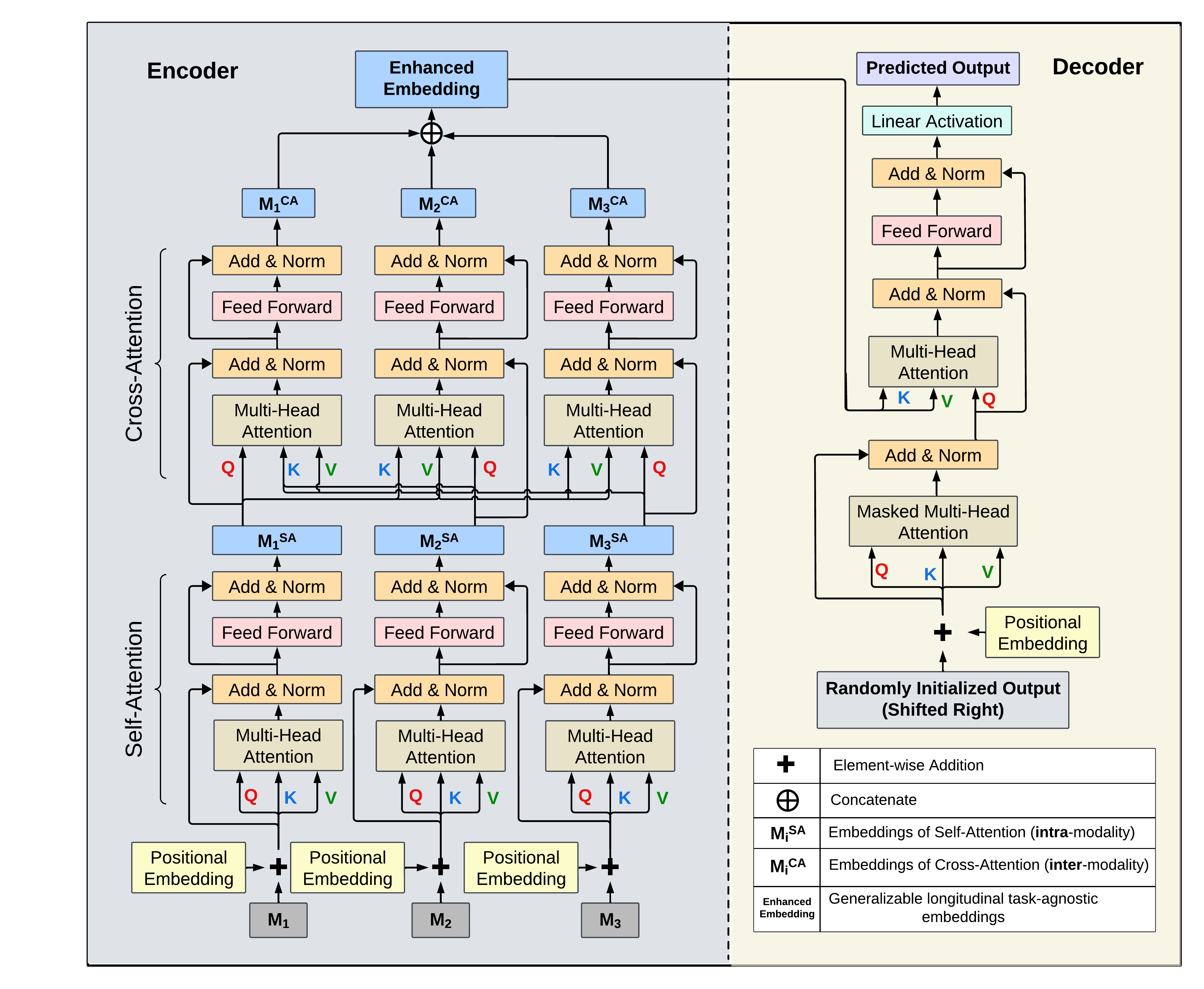}
    \EndAccSupp{}
    \caption{The architecture of CAAT-EHR.}
    \label{fig:CAAT-EHR}
\end{figure}

Figure~\ref{fig:CAAT-EHR} illustrates the architecture of CAAT-EHR, highlighting the encoder and decoder components and their interactions. This architecture is used during the pre-training phase to learn task-agnostic longitudinal embeddings in a self-supervised manner, as depicted in the pre-training step of Figure~\ref{fig:processing}B. The encoder generates embeddings by leveraging self- and cross-attention mechanisms, while the decoder serves as a supervision mechanism to optimize the encoder's output through autoregressive modeling.


\subsubsection{The Encoder}
The encoder processes three input modalities, $M_1$, $M_2$, and $M_3$, representing continuous, categorical, and textual clinical notes data for MIMIC-III, or cognitive measurement, MRI, and DNA methylation data for ADNI. Initially, positional encoding is applied to each data modality to incorporate sequence order information, as described in the original Transformer architecture \cite{vaswani2017attention}.
To capture intra-modality temporal dependencies, each data modality $M_m \in \mathbb{R}^{T \times d_m}$ is processed independently by a multi-head self-attention encoder. For self-attention, query $Q$, key $K$, and value $V$ are identical ($Q$ = $K$ = $V$ = $M_m$). The scaled dot-product attention is defined as (Equation~\ref{eq:1}).
\begin{equation}
    \text{Attention}(Q, K, V) = \text{Softmax}\left( \frac{QK^\top}{\sqrt{d_k}} \right)V
\label{eq:1}
\end{equation}

In our implementation, the Keras MultiHeadAttention layer internally applies learnable linear projections to obtain per-head queries, keys, and values. We do not apply an additional modality-specific projection layer prior to attention, which reduces architectural complexity and avoids introducing extra trainable parameters beyond those inherent to the attention mechanism. Multi-head attention is computed as:

\begin{equation}
\text{MultiHead}(Q, K, V) = \text{Concat}(\text{head}_1, \ldots, \text{head}_h) W^O
\label{eq:2}
\end{equation}

\begin{equation}
\text{head}_j = \text{Attention}(Q W_j^Q, K W_j^K, V W_j^V)
\label{eq:3}
\end{equation}

where $W_j^Q \in \mathbb{R}^{d_{\text{m}} \times d_k}$, 
$W_j^K \in \mathbb{R}^{d_{\text{m}} \times d_k}$, 
$W_j^V \in \mathbb{R}^{d_{\text{m}} \times d_v}$, and 
$W^O$ projects the concatenated head outputs back to the modality feature dimension. The per-head key dimension 
$d_k$ (and value dimension $d_v$) is specified by the attention layer hyperparameters.

The self-attention outputs are denoted as $M_1^{SA}$, $M_2^{SA}$, and $M_3^{SA}$ for the respective modalities. Residual connections and layer normalization are applied after both the attention and feed-forward (FFN) sublayers to stabilize training and preserve temporal feature representations, as proposed in the original Transformer architecture \cite{vaswani2017attention}.

Following self-attention, multi-head attention layers are used to perform cross-attention (Equation~\ref{eq:1}) between each pair of data modality. In this cross-attention step, $Q$ is derived from self-attention output $M_i^{{SA}}$ of modality $i$, while $K$ and $V$ are derived from the self-attention output $M_j^{{SA}}$ of another modality $j$, where \begin{math} i, j \in \{1,2,3\}\end{math}, and $i \ne j$. This design allows each modality to attend to complementary temporal information from the other modalities.

The dimensions of $Q$, $K$, and $V$ are now determined by the embedding size of the corresponding self-attention output. As in self-attention, the Keras MultiHeadAttention layer internally applies learnable linear projections to obtain per-head queries, keys, and values, and the concatenated head outputs are projected back to the modality feature dimension by the output projection matrix. Each cross-attention output $M_i^{CA}$ is computed by applying dropout, followed by residual connection and layer normalization to the intermediate cross-attention result obtained when modality $i$ attends to the other modalities. This preserves the original modality representation while incorporating inter-modality contextual information.
Like self-attention, the outputs from all attention heads in cross-attention are concatenated and linearly transformed using a trainable weight matrix, as described in Equations (~\ref{eq:2}) and (\ref{eq:3}). In our implementation, the per-head key and value dimensions are specified by the attention layer hyperparameters, rather than being strictly tied to the embedding dimension.

Finally, the cross-attention outputs, $M_1^{CA}$, $M_2^{CA}$ and $M_3^{CA}$, are concatenated to form a single representation representing the task-agnostic longitudinal embeddings. This representation serves as the encoder’s output, which is further optimized by the decoder during pre-training. Once pre-training is complete, the encoder is utilized to generate task-agnostic longitudinal embeddings for any downstream task data, enabling robust performance across various downstream tasks.

\subsubsection{The Decoder}
  The decoder serves solely as a supervision mechanism to refine the encoder’s output by autoregressively predicting data for the next two time points in the input sequence, mimicking the next-word prediction task in NLP. The decoder predicts the next two time points to balance capturing temporal dependencies and maintaining model stability, as predicting more than two time points risks compounding errors and increased optimization complexity.

The decoder starts with randomly initialized values representing its output. Positional encoding is applied to this initial output to incorporate temporal information. The initial output is then passed through a masked multi-head attention layer, which functions similarly to the self-attention mechanism in the encoder (Equation~\ref{eq:1}) but operates in an autoregressive manner. Masking is applied to the output sequence (shifting it to the right) to ensure that the prediction for each time point only depends on the preceding time points. In this masked attention mechanism, the $Q$, $K$, and $V$ are derived from the same data (i.e., the decoder's current output).

Next, the output of the masked multi-head attention layer is passed into another multi-head attention layer, which applies a cross-attention mechanism between the current output of the masked attention layer and the encoder’s output. This step aligns the generated output with the relevant encoded information. Here, $Q$ is derived from the decoder's output, while $K$ and $V$ are derived from the encoder's output (Equation~\ref{eq:1}).

Finally, the decoder generates output that represents the data for the next two time points. The primary objective of the entire model is to minimize the discrepancy between the decoder’s generated output and the actual target data (i.e., the data for the last two time points). This alignment is achieved by reducing the error, specifically using the Mean Squared Error(MSE) loss, which is computed as shown in Equation~\ref{eq:4}.
\begin{equation}
\text{MSE} = \frac{1}{N_{emb}} \sum_{i=1}^{N_{emb}} (z_i - \hat{z}_i)^2
\label{eq:4}
\end{equation}

where $z_i$ and $\hat{z}_i$ represent the actual and predicted values for the $i$-th sample, respectively, and $N_{emb}$ is the total number of samples (e.g., patients) in the embedding task dataset.

For the optimization process, CAAT-EHR was pre-trained using the Adaptive Moment Estimation (Adam) optimizer \cite{kingma2014adam}. Hyperparameters were tuned using 10\% of the embedding task data as validation data (Supplemental Table 5).


\subsection{Experimental Setup}
\label{subsec: 2.3}
After pre-training the proposed CAAT-EHR model on the embedding task data, the pre-trained CAAT-EHR encoder was retained to generate task-agnostic longitudinal embeddings of the downstream task data. Using this new representation, we trained models for various tasks (see Section 2.1.5).


To evaluate the quality of the generated embeddings, we compared the performance of predictive models trained on three types of input: (1) the embeddings produced by CAAT-EHR, (2) the original raw downstream task data, and (3) the embeddings generated by the LSTM-AE encoder after it was trained to reconstruct the embedding task data. We evaluated TA-RNN, LSTM, RF, and SVM using the raw downstream task data and the embeddings generated by CAAT-EHR and LSTM-AE. RF and SVM were evaluated on the aggregated data only, as they cannot process longitudinal sequences.

In addition, we included BEHRT and OvO as comparative baselines. BEHRT was pre-trained on the embedding task data and fine-tuned on the raw downstream task data for the specific prediction tasks as it does not generate embeddings and operates directly on raw inputs. On the other hand, OvO was trained and evaluated solely on the raw downstream task data, as it employs a cross-attention mechanism to integrate multimodal inputs end-to-end, without generating or utilizing intermediate embedding representations.


All models were evaluated using repeated stratified train/test splits. For the MIMIC-III dataset, to prevent data leakage, no ICU stays from the same patient were allowed to appear in both the training and testing sets.  In each split, 70\% of the data was used for training and 30\% for testing. This procedure was repeated across five random seeds, and each split was run 10 times (50 runs total). Results are reported as the mean and standard error across runs. Evaluation metrics include F1 score and Area Under the Receiver Operating Characteristic Curve (AUC).

Hyperparameters for TA-RNN, LSTM, OvO, RF, and SVM were tuned using 5-fold cross-validation (Supplemental Tables 6–10), while hyperparameters for LSTM-AE and BEHRT were tuned using 10\% of the embedding task dataset as validation data (Supplemental Tables 11 and 12, respectively).


\section{Results and Discussion}
We evaluated CAAT-EHR on three downstream tasks, with performance measured by F1 score and AUC and statistical significance assessed using paired two-sided t-tests.



\subsection{Evaluation of CAAT-EHR Using the MIMIC-III Data}
\label{subsec:3.1}



Table~\ref{tab:1} presents the average F1 score and AUC with the standard error of each model for the length-of-stay prediction task. For models evaluated on both raw and CAAT-EHR embeddings, CAAT-EHR embeddings led to improved performances in three out of four models for both F1 score and AUC, demonstrating their generalizability and utility for both temporal and non-temporal modeling tasks in ICU length-of-stay prediction. Specifically, the results demonstrate that LSTM trained on CAAT-EHR embeddings achieved the highest overall performance, in terms of F1 score and AUC, highlighting the strength of CAAT-EHR in capturing temporal patterns relevant to length of stay in ICU. On the other hand, when LSTM trained on raw EHR data, F1 score and AUC dropped significantly by 16.59\% (p-value = 2.5e-08) and 13.77\% (p-value = 1.7e-09), respectively. 

Among the standalone methods, CAAT-EHR embeddings significantly outperformed OvO (p-value = 0.0005 for F1 and p-value = 0.001 for AUC) and BEHRT (p-value = 6.4e-12 for F1 and p-value = 1.8e-19 for AUC). Since BEHRT was originally designed for textual data, its low performance could be due to its modality-specific nature.

\begin{table}
  \caption{F1 score and AUC for TA-RNN, LSTM, OvO, BEHRT, RF, and SVM models trained on the MIMIC-III downstream task data for ICU length-of-stay prediction, based on different embedding approaches. Results are reported as mean ± standard error. \underline{Underlined} values indicate the highest performance across all models for each metric, while bold values indicate the highest performance within each model and metric group. N/A: Using raw data without any embedding.}
  \label{tab:1}
   \centering
  \begin{tabular}{llcc}
    \toprule
    \textbf{Model} & \textbf{Embedding} & \textbf{F1} & \textbf{AUC} \\
    \midrule
    \multirow{3}{*}{TA-RNN} 
        & CAAT-EHR   & \textbf{0.529$\pm$0.016} & \textbf{0.586 $\pm$0.015} \\
        & LSTM-AE    & 0.479$\pm$0.007 & 0.563 $\pm$0.005 \\
        & N/A        & 0.444$\pm$0.013 & 0.519$\pm$0.006 \\
    \midrule
    \multirow{3}{*}{LSTM} 
        & CAAT-EHR   & \textbf{\underline{0.657$\pm$0.013}} & \textbf{\underline{0.704$\pm$0.011}} \\
        & LSTM-AE    & 0.401$\pm$0.016 & 0.515$\pm$0.004 \\
        & N/A        & 0.548$\pm$0.016 & 0.607$\pm$0.010 \\
    \midrule
    OvO & N/A        & 0.625$\pm$0.007 & 0.635$\pm$0.011\\
    \midrule
    BEHRT & N/A        & 0.453$\pm$0.014 & 0.520$\pm$0.003 \\
    \midrule
    \multirow{3}{*}{RF} 
        & CAAT-EHR   & 0.609$\pm$0.005 & 0.598$\pm$0.004 \\
        & LSTM-AE    & 0.446$\pm$0.001 & 0.503$\pm$0.001 \\
        & N/A        & \textbf{0.650$\pm$0.002} & \textbf{0.630$\pm$0.002} \\
    \midrule
    \multirow{3}{*}{SVM} 
        & CAAT-EHR   & \textbf{0.642$\pm$0.003} & \textbf{0.625$\pm$0.002} \\
        & LSTM-AE    & 0.447$\pm$0.002 & 0.502$\pm$0.0008 \\
        & N/A        & 0.437$\pm$7.8e-18 & 0.500$\pm$0.000 \\
    \bottomrule
  \end{tabular}
\end{table}




\begin{table}
  \caption{F1 score and AUC for TA-RNN, LSTM, OvO, BEHRT, RF, and SVM models trained on the MIMIC-III downstream task data for mortality prediction, based on different embedding approaches. Results are reported as mean ± standard error. \underline{Underlined} values indicate the highest performance across all models for each metric, while bold values indicate the highest performance within each model and metric group. N/A: Using raw data without any embedding.}
  \label{tab:2}
   \centering
  \begin{tabular}{llcc}
    \toprule
    \textbf{Model} & \textbf{Embedding} & \textbf{F1} & \textbf{AUC} \\
    \midrule
    \multirow{3}{*}{TA-RNN} 
        & CAAT-EHR   & \textbf{0.583$\pm$0.009} & \textbf{0.671$\pm$0.012} \\
        & LSTM-AE    & 0.468$\pm$0.009 & 0.551$\pm$0.004 \\
        & N/A        & 0.433$\pm$0.017 & 0.532$\pm$0.007 \\
    \midrule
    \multirow{3}{*}{LSTM} 
        & CAAT-EHR   & \textbf{0.607$\pm$0.010} & \textbf{\underline{0.691$\pm$0.013}} \\
        & LSTM-AE    & 0.310$\pm$0.024 & 0.501$\pm$0.001 \\
        & N/A        & 0.538$\pm$0.013 & 0.623$\pm$0.011 \\
    \midrule
    OvO & N/A        & 0.634$\pm$0.001 & 0.645$\pm$0.006 \\
    \midrule
    BEHRT & N/A        & 0.468$\pm$0.017 & 0.519$\pm$0.002 \\
    \midrule
    \multirow{3}{*}{RF} 
        & CAAT-EHR   & \textbf{0.609$\pm$0.004} & \textbf{0.590$\pm$0.003} \\
        & LSTM-AE    & 0.526$\pm$0.002 & 0.530$\pm$0.001 \\
        & N/A        & 0.607$\pm$0.003 & 0.585$\pm$0.002 \\
    \midrule
    \multirow{3}{*}{SVM} 
        & CAAT-EHR   & \textbf{\underline{0.647$\pm$0.003}} & \textbf{0.628$\pm$0.002} \\
        & LSTM-AE    & 0.496$\pm$0.003 & 0.513$\pm$0.001 \\
        & N/A        & 0.497$\pm$0.002 & 0.516$\pm$0.001 \\
    \bottomrule
  \end{tabular}
\end{table}
 The results presented in Table~\ref{tab:2} show that CAAT-EHR provides robust embeddings for in-hospital mortality prediction. LSTM models trained on CAAT-EHR embeddings significantly outperformed LSTM trained on raw data, with p-value = 1.0e-05 for F1 score and p-value = 1.2e-05 for AUC. Similarly, RF and SVM models trained using embeddings generated by CAAT-EHR  had significantly higher F1 and AUC than the other RF and SVM models (p-value $<$ 1e-3 for F1 and AUC). Among the standalone models, OvO showed comparable performance for F1, but underperformed for AUC (p-value = 0.0003), while BEHRT performed poorly (p-value = 5.9e-09 for F1 and p-value = 1.2e-16 for AUC), reflecting its limited applicability to structured clinical data. 



\subsection{Evaluation of CAAT-EHR Using the ADNI Data}
We also evaluated CAAT-EHR for the AD prediction task using the ADNI data (Table~\ref{tab:3}). For models evaluated on both raw data and CAAT-EHR embeddings, CAAT-EHR embeddings led to improved performances in all four models for both F1 score and AUC, indicating the effectiveness of CAAT-EHR. For instance, LSTM models trained on CAAT-EHR embeddings significantly outperformed those trained on LSTM-AE embeddings (p-value = 1.2e-09 for F1 and p-value = 1.9e-11 for AUC), demonstrating CAAT-EHR’s strength in capturing temporal patterns. 






Additionally, non-temporal models such as RF and SVM performed significantly better with aggregated CAAT-EHR embeddings than the counterpart embedding methods, achieving higher F1 scores and AUC (p-value $<$ 1e-9). Among the standalone methods, BEHRT and OvO exhibited significantly lower performances (p-value $\leq$  3.3e-06 for BEHRT and p-value $\leq$ 0.044 for OvO). 


Among all models, TA-RNN trained on CAAT-EHR embeddings achieved the highest F1 score and AUC. These findings underscore the utility of CAAT-EHR embeddings for both temporal and aggregated analyses, particularly for the models that do not have internal embedding generation step. 



\begin{table}
  \caption{F1 score and AUC for TA-RNN, LSTM, OvO, BEHRT, RF, and SVM models trained on the ADNI downstream task data for AD prediction, based on different embedding approaches. Results are reported as mean ± standard error. \underline{Underlined} values indicate the highest performance across all models for each metric, while bold values indicate the highest performance within each model and metric group. N/A: Using raw data without any embedding.}
  \label{tab:3}
  \centering
  \begin{tabular}{llcc}
    \toprule
    \textbf{Model} & \textbf{Embedding} & \textbf{F1} & \textbf{AUC} \\
    \midrule
    \multirow{3}{*}{TA-RNN} 
        & CAAT-EHR   & \textbf{\underline{0.869$\pm$0.002}} & \textbf{\underline{0.876$\pm$0.002}}
 \\
        & LSTM-AE    & 0.749$\pm$0.008 & 0.761$\pm$0.007 \\
        & N/A        & 0.854$\pm$0.010 & 0.859$\pm$0.008 \\
    \midrule
    \multirow{3}{*}{LSTM} 
        & CAAT-EHR   & \textbf{0.861}$\pm$\textbf{0.003} & \textbf{0.865}$\pm$\textbf{0.003} \\
        & LSTM-AE    & 0.786$\pm$0.009 & 0.799$\pm$0.006 \\
        & N/A        & 0.846$\pm$0.010 & 0.853$\pm$0.007 \\
        \midrule
    OvO & N/A        & 0.807$\pm$0.028 & 0.836$\pm$0.019 \\
\midrule
    BEHRT & N/A        & 0.756$\pm$0.022 & 0.791$\pm$0.016 \\
    \midrule
    \multirow{3}{*}{RF} 
        & CAAT-EHR   & \textbf{0.818$\pm$0.003} & \textbf{0.823 $\pm$0.002} \\
        & LSTM-AE    & 0.681$\pm$0.009 & 0.689$\pm$0.008 \\
        & N/A        & 0.813$\pm$0.003 & 0.818$\pm$0.003 \\
    \midrule
    \multirow{3}{*}{SVM} 
        & CAAT-EHR   & \textbf{0.856$\pm$0.002} & \textbf{0.857$\pm$0.002} \\
        & LSTM-AE    & 0.706$\pm$0.016 & 0.715$\pm$0.012 \\
        & N/A        & 0.689$\pm$0.019 & 0.710$\pm$0.016 \\
    \bottomrule
  \end{tabular}
\end{table}


\subsection{Ablation Study}

We conducted ablation studies to determine the contributions of cross-attention, the autoregressive decoder, and individual data modalities. Table~\ref{tab:4} presents the F1 and AUC scores for all prediction tasks. These metrics were calculated using embeddings generated by the full CAAT-EHR model and five ablated versions: (1) without the cross-attention component; (2) without the autoregressive component where the decoder was pre-trained based on reconstructing the input sequence instead of predicting the next two time points; and (3–5) individual modality pre-training, including continuous, categorical, and clinical notes data for MIMIC-III, and cognitive, MRI, and DNA methylation data for ADNI. To assess the quality of embeddings across tasks, LSTM models were trained on the embeddings produced by each version of the model.

\begin{table}
\small
  \caption{F1 score and AUC for different prediction tasks using the proposed CAAT-EHR model and its ablated versions. \underline{Full:} the complete proposed model, \underline{Without CA:} the proposed model without the cross-attention component, \underline{Without AR:} the proposed model without the autoregressive component, replaced with reconstruction-based pre-training, \underline{Only X:} the proposed model pre-trained only on data modality X. CONT: Continuous data, CATE: Categorical data, COG: Cognitive score data, MRI: MRI data, NOTE: Clinical Notes, DM: DNA Methylation data. Best values for each prediction task are shown in bold.}
  \label{tab:4}
  \centering
  \begin{tabular}{llcc}
    \toprule
    \textbf{Prediction Task} & \textbf{Model Version} & \textbf{F1} & \textbf{AUC} \\
    \midrule
    \multirow{3}{*}{Length of stay}
        & Full         & \textbf{0.657$\pm$0.013} & \textbf{0.704$\pm$0.011} \\
        & Without CA   & 0.627$\pm$0.015 & 0.665$\pm$0.014 \\
        & Without AR   & 0.624$\pm$0.016 & 0.670$\pm$0.014 \\
        & Only CONT   & 0.499$\pm$0.007 & 0.548$\pm$0.004 \\
        & Only CATE   & 0.656$\pm$0.016 & 0.693$\pm$0.014 \\
        & Only NOTE   & 0.500$\pm$0.008 & 0.572$\pm$0.004 \\
    \midrule
    \multirow{3}{*}{Mortality}
        & Full         & \textbf{0.607$\pm$0.010} & \textbf{0.691$\pm$0.013} \\
        & Without CA   & 0.574$\pm$0.014 & 0.658$\pm$0.012 \\
        & Without AR   & 0.586$\pm$0.013 & 0.664$\pm$0.012 \\
        & Only CONT   & 0.500$\pm$0.006 & 0.562$\pm$0.005 \\
        & Only CATE   & 0.606$\pm$0.010 & 0.683$\pm$0.012 \\
        & Only NOTE   & 0.407$\pm$0.013 & 0.533$\pm$0.003 \\
    \midrule
    \multirow{3}{*}{AD Diagnosis}
        & Full         & \textbf{0.861}$\pm$\textbf{0.003} & \textbf{0.865}$\pm$\textbf{0.003} \\
        & Without CA   & 0.845$\pm$0.003 & 0.851$\pm$0.002 \\
        & Without AR   & 0.826$\pm$0.005 & 0.836$\pm$0.004 \\
        & Only COG   & 0.860$\pm$0.003 & 0.864$\pm$0.002 \\
        & Only MRI   & 0.781$\pm$0.005 & 0.788$\pm$0.004 \\
        & Only DM   & 0.501$\pm$0.006 & 0.541$\pm$0.003 \\
    \bottomrule
  \end{tabular}
\end{table}

In length-of-stay prediction task, removing the cross-attention component worsened the results significantly (p-value = 0.027 for F1 and p-value = 0.001 for AUC). Removing the autoregressive component also resulted in a significant drop in performance (p-value = 0.045 for F1 and p-value = 0.019 for AUC), highlighting the importance of the autoregressive component. Pre-training each modality independently performed worse relative to the proposed multimodal CAAT-EHR model (Full), with statistically significant differences (p-value $<$ 0.05) for continuous and clinical notes modalities.  


For mortality prediction, excluding the cross-attention component caused a significant decrease in performance (p-value = 0.017 for F1 and p-value = 0.025 for AUC) and removing the autoregressive component showed a trend towards reduced performance with marginal significance (p-value = 0.098 for F1 and p-value = 0.065 for AUC). This highlights the importance of both components. Similar to length-of-say prediction, the models pre-trained on individual modalities performed worse than the multimodal model (Full), with statistically significant gaps (p-value $<$ 0.05) for the continuous and clinical notes modalities. On the other hand, for both length-of-stay and mortality prediction tasks, the models trained using only the embedding of the categorical data modality achieved near-optimal performance, highlighting the importance of this data modality. 

For AD prediction, the full CAAT-EHR model outperformed the other versions, achieving the highest F1 score and AUC. Removing both the cross-attention component and the autoregressive component caused a significant decrease in performance (p-value $\leq 0.0005$ for F1 and p-value $\leq 0.0002$ for AUC), highlighting the importance of both components. Pre-training on individual modalities resulted in inferior performance compared to the proposed multimodal model (Full), with statistically significant differences (p $<$ 0.05) for MRI and DNA methylation. The model using only cognitive-score data performed slightly worse than the multimodal model, likely because cognitive tests inform clinical AD diagnosis.


\section{Conclusion}

In this study, we introduced CAAT-EHR, a Cross-Attentional Autoregressive Transformer architecture designed to generate task-agnostic embeddings of multimodal longitudinal EHR data. CAAT-EHR effectively integrates temporal, contextual, and multimodal relationships. Using benchmark datasets (MIMIC-III and ADNI), we demonstrated that models trained on CAAT-EHR-generated embeddings outperformed those trained on raw EHR data and embeddings generated by baseline methods across various downstream tasks, including mortality prediction, ICU length-of-stay estimation, and AD diagnosis prediction. Ablation studies highlighted that multimodal data, cross-attention fusion, and the autoregressive decoder are critical for effective representation learning. Future work could explore pre-training on larger datasets and expanding the model's applications to other healthcare challenges. Although CAAT-EHR currently integrates three data modalities for multimodal EHR representation learning, it can be seamlessly extended to incorporate additional modalities.








\section{Acknowledgments}
This research was supported by the National Institute of General Medical Sciences of the National Institutes of Health under Award Number R35GM133657. Additional support was provided through a summer assistantship from the BioDiscovery Institute and startup funds from the University of North Texas. Data used in preparation of this article were obtained from the ADNI database (adni.loni.usc.edu). As such, the investigators within the ADNI contributed to the design and implementation of ADNI and/or provided data but did not participate in analysis or writing of this report. A complete listing of ADNI investigators can be found at: https://adni.loni.usc.edu/wp-content/uploads/how\_\allowbreak to\_\allowbreak apply/ADNI\_\allowbreak Acknowledgement\_\allowbreak List.pdf.

\section{Competing interests}
No competing interest is declared.

\section{Data availability}
This study uses publicly available data from MIMIC-III (https://physionet.org/content/mimiciii/1.4/) and ADNI (https://adni.loni.usc.edu/).

\section{Author contributions statement}
Conceptualization: M.A.O, S.B.; Methodology: M.A.O,  S.B., S.C.; Data Collection: M.A.O, S.C.; Running experiments: M.A.O, S.C.; Writing - Original Draft: M.A.O, S.B.; Writing - Review \& Editing: S.C., S.B., M.A.O.; Visualization: M.A.O., S.C.; Supervision: S.B.



\end{document}